\documentclass[letterpaper, 10 pt, conference]{ieeeconf}  

\IEEEoverridecommandlockouts                              
\usepackage[utf8]{inputenc}
\usepackage[T1]{fontenc}
\usepackage{amsmath,amsfonts,amssymb}

\usepackage{graphics} 
\usepackage{epsfig} 
\usepackage{mathptmx} 
\usepackage{mathptmx} 
\usepackage{amsmath} 
\usepackage{amssymb}  
\usepackage{hyperref}
\usepackage{adjustbox}
\newcommand{\ts}{\textsuperscript}

\title{\LARGE \bf Weakly supervised deep learning-based intracranial hemorrhage localization}

\author{Jakub Nemcek$^{1}$, Tomas Vicar$^{1}$ and Roman Jakubicek$^{1}$
\thanks{$^{1}$J. Nemcek, T. Vicar, R. Jakubicek  are from Department of Biomedical Engineering, Faculty of Electrical Engineering and Communications, Brno University of Technology, Brno, Czech Republic.      Contact mail: {\tt\small nemcek@vut.cz}}%
}

\begin{document}

\maketitle
\thispagestyle{empty}
\pagestyle{empty}

\begin{abstract}
Intracranial hemorrhage is a life-threatening disease, which requires fast medical intervention. Owing to the duration of data annotation, head CT images are usually available only with slice-level labeling. This paper presents a weakly supervised method of precise hemorrhage localization in axial slices using only position-free labels, which is based on multiple instance learning. An algorithm is introduced that generates hemorrhage likelihood maps and finds the coordinates of bleeding. The Dice coefficient of 58.08 \% is achieved on data from a publicly available dataset.    

\end{abstract}

\section{INTRODUCTION}

Intracranial hemorrhage (ICH) is a relatively common life-threatening disease (25 per 100,000 people per year) that may develop after physical trauma or non-traumatically. 
The significance of this event is given by a high 30-days mortality rate (up to 52 \%) and a large risk of lasting consequences among survivors \cite{Caceres2012}. For these reasons, the fast discovery of the disease is crucial for the early initiation of treatment. If the diagnosis of the disease is delayed (within minutes), it increases the risk of permanent brain dis-function, or it can be even fatal. Modern brain CT analysing computer systems can provide fast and effective support for computer-aided diagnosis and can therefore be very useful for physicians' decisions, especially in acute cases. 

Nowadays, most state-of-the-art methods are focused on deep learning approaches, especially using convolutional neural networks (CNN) and their modifications or combinations. Published detection algorithms \cite{reserse_8} and \cite{reserse_9} use 3D CNN-based classification on the patient level that provide a decision about the presence of ICHs in patient scans. A combination of 2D CNN and LSTM (Long short-term memory) algorithm for ICH detection in CT slices was designed by the authors of \cite{reserse_13}. Another combination of CNN and recurrent neural network was published in \cite{reserse_11} that includes classification into its sub-types.

The algorithm for 2D ICH segmentation including its type classification based on cascade CNN model was applied by the authors of \cite{reserse_6}. For the same task, the authors of \cite{reserse_7} suggested a hybrid 2D/3D approach using Mask Regional-CNN algorithm. 
The well-known U-net architecture of CNN has been used in \cite{reserse_5} for 2D ICH segmentation. A similar 3D approach was introduced by the authors of \cite{reserse_10}.

One of the first published mentions of the possible utilizing of attention maps for ICH detection or segmentation is in \cite{reserse_3}. Here, the authors cursorily validated the ICHs center detection via plain thresholding of the attention maps (as an appendix of their manuscript). In \cite{reserse_11}, the attention maps were displayed for mere visualization of the network field of view.


The presented approach of ICH localization is based on the detection of local extrema in attention maps, which can be understood as the likelihood of ICH occurrence for each pixel. These maps are produced by the proposed weakly supervised approach based on Multiple Instance Learning (MIL) \cite{Attention}. Its advantage is a position-free learning, thus precise position annotations are not needed for training, and slice-level annotations (healthy/ICH) are quite sufficient. In addition, unlike \cite{reserse_3}, we apply a special local maxima detector \cite{Koenderink1992} to the attention map that directly leads to the precise localization of ICHs.


\section{METHODS}

\subsection{Experimental data}
The head CT data from two publicly available datasets were used in this work. More than 25,000 non-contrast brain CT scans manually labeled for each slice have been released for the purposes of a challenge held by the Radiological Society of North America (RSNA) \cite{Flanders20200501}. Besides, almost 500 head scans are available within the CQ500 dataset \cite{HeadStudy2018} together with patient-level annotations of different pathologies, including ICHs. For the purposes of this work, CT images with soft-tissue reconstruction kernel including ICHs or healthy patient scans are used (other pathologies are included only in case of co-occurrence with ICH). An extension of CQ500 annotation has been made public on the PhysioNet \cite{PhysioNet} in the BHX dataset \cite{BHX}. The authors released bounding boxes (BBs) labeling the position of ICHs, which are gained either manually by three radiologists for slices with the lower axial resolution or by an extrapolation. The reason that CQ500 is used is that the ICHs positional labels are required for the evaluation of the proposed method. Moreover, the evaluation on the dataset completely different from the training and validation set proves the generalization of the proposed approach to new data.

\begin{figure*}	[!tb]
			\centering
			\includegraphics[width=0.95\linewidth]{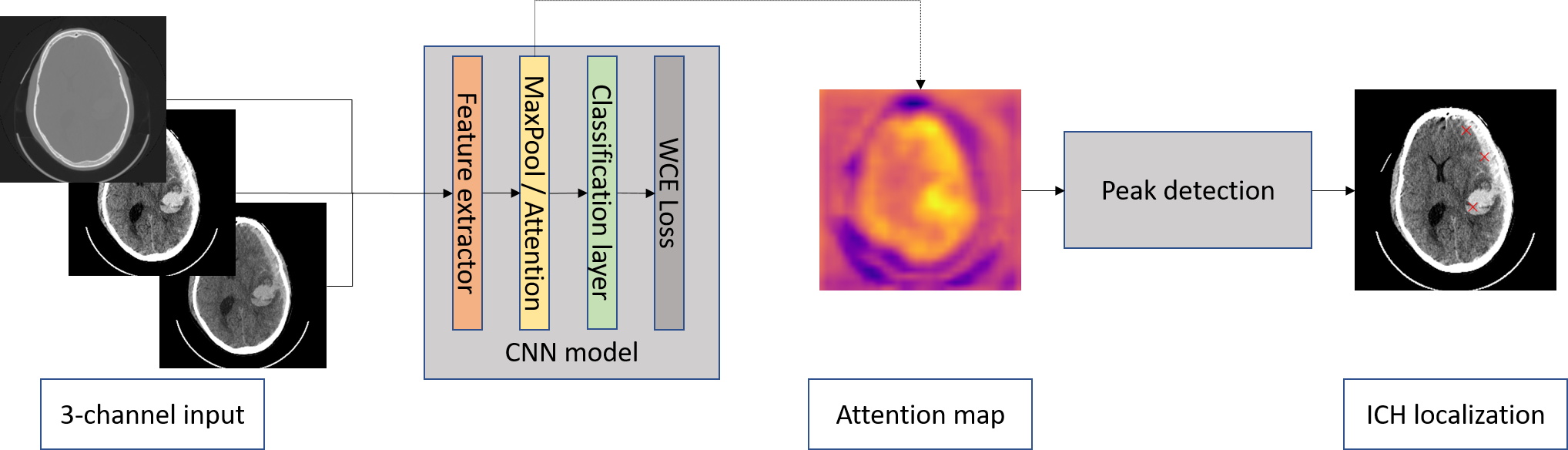}
			\caption{The overall pipeline of weakly supervised ICH detection algorithm.}
		\label{fig:pipeline}
\end{figure*}

\subsection{Weakly supervised detection}
The aim of this work is to build a CNN-based detector that can accurately localize ICHs in axial CT slices, using only slice-level annotations. A classification CNN model is trained to predict both the probability that ICH is present in the image and an attention map providing information about the likelihood of the area being affected. Using this heat map, a detector is proposed that provides exact positional information about the location of ICHs. The block scheme of the algorithm is shown in Fig~\ref{fig:pipeline}. 

To form an input of the model, the axial slices from CT scans are taken as the first channel. To form the other two channels, the contrast enhancement is applied with respect to two radiological windows, i.e., brain (L=40, W=80) and subdural (L=50, W=130), which has been proven to be effective in our previous work \cite{Nemcek20211130}. Besides, all data are standardized by the mean and standard deviation of the whole dataset.

ResNet-like architecture \cite{He2016} is chosen for the classification CNN. The feature extractor has a total of 18 convolution layers and undersamples the feature maps to a quarter of the original size. The resulting feature map is input into either the max-pooling or attention layer. Both layers result in creating outputs (for details see section \ref{layers}) that are processed by the fully-connected layer providing a binary classification (pathological or non-pathological slices). In parallel, either the activation maps ahead of global max-pooling or the weights of the attention layer are resized to the size of the original image to form the attention maps, which are further processed by the proposed ICH detector described in the section \ref{detector}. 

To obtain the desired results, the CNN classifier is trained using CT scans from the RSNA dataset and a weighted cross-entropy loss function. The data was divided in proportion 97:3 into training and validation sets.


\subsection{Max-pooling and attention layer}
\label{layers}

In the case of max-pooling, the forward propagation can be described by:
\begin{equation}
    z = \max_{k=1,...,K}(h_{k}) 
\end{equation}
where $z$ denotes the value of the output and $h_{k}$ is the value of the feature image on the network output, where in our case, global pooling over all positions $k$ in the image is applied. By taking the maximal value from the image, the output captures the strongest feature position $h_{k}$. Feature image $h_{k}$ shows the prediction of ICH at individual positions, and thus, it can be used as an attention map. However, it is limited to the scalar feature map $h_{k}$, which limits the classification capability of the network \cite{Attention}. 

The MIL attention layer \cite{Attention} computes the value of the output feature vector $\textbf{z}$ as a weighted average of low-dimensional embeddings $\textbf{h}_{k}$:
\begin{equation}
    \textbf{z} = \sum_{k=1}^{K} a_k \textbf{h}_k.
\end{equation}
The weights $a_k$ are determined by a two one-layer fully connected neural networks applied to each position of the feature image, where outputs are element-wise multiplied and transformed with softmax to achieve that $\sum_k a_k = 1$: 
\begin{equation}
    a_k = \frac{\exp\{{\textbf{w}^\intercal(tanh(\textbf{Vh}_k^\intercal) \odot sigm(\textbf{Uh}_k^\intercal) )}\}}
    {\sum_{j=1}^{K} \exp\{{\textbf{w}^\intercal(tanh(\textbf{Vh}_j^\intercal) \odot sigm(\textbf{Uh}_j^\intercal) )}\}},
\end{equation}
where $\textbf{w} \in \mathbb{R}^{L\times1}$, $\textbf{V} \in \mathbb{R}^{L\times M}$ and $\textbf{U} \in \mathbb{R}^{L\times M}$ are parameters of the small neural network, $\odot$ is an element-wise multiplication, $tanh(\dot)$ and $sigm(\dot)$ are activation functions representing hyperbolic tangent and sigmoid nonlinearity. In the case of image, the fully connected neural network can be realized by 1x1 convolution. The attention map $a_k$ represents the importance of individual positions to form the network's decision \cite{Attention}. 

\subsection{Detector}
\label{detector}
Generated attention maps mark ICHs as the areas with higher pixel values in comparison to the background. Hence, the detector finds local maxima via comparison of the original image with a grayscale-dilated (via the maximum filter) image \cite{Koenderink1992}, and returns their coordinates representing individual bleedings. To suppress irrelevant peaks with peak prominence smaller than $h$, h-maxima transform \cite{thirusittampalam2013novel} is performed. Besides, to avoid multiple and false findings, the value of local maxima must be higher than the threshold $T$ and the minimal allowed distance between peaks is $d$. The optimal detector's parameters $h$, $T$ and $d$ were found by Bayesian optimization.

\subsection{Implementation details}

Network was trained using Adam optimizer \cite{kingma2014adam} with weight decay of $10^{-6}$, 1\ts{st} and 2\ts{nd} moment estimates were set to 0.9 and 0.999, respectively. Learning rate 0.01 decaying to 10 \% after 20, 10 and 5 epochs was used. Network from epoch with optimal accuracy on validation set was used to generate results. Batch size 128 was used, where images were randomly cropped to size 256. Furthermore, augmentation was performed using: random mirroring, random affine transformation (30 deg rotation, max 10\% scaling, max 5\% shearing), brightness multiplying (max 1.02x), brightness addition (max $\pm$0.2), blurring/sharpening (max 0.5 Gaussian sigma and max subtraction of 0.5 Laplacian). The network contains 18 convolutional layers with 3 levels separated by pooling layers, where each level contains 3 residual blocks \cite{He2016}.  Code is available at \url{https://github.com/tomasvicar/ICH-MIL-attention-based-detector}.

Parameters of the detector were determined by Bayesian optimization \cite{snoek2012practical} (implementation from \cite{bayes_opt}) maximizing Dice coefficient (DC) while using 2/5 of CT data from the CQ500 dataset. For the attention map given by max-pooling layer and attention layer, the optimal parameters are: $h=0.024$, $T=0.76$, $d=10$, and $h=0.0038$, $T=0.024$, $d=58$, respectively.

\begin{figure*}	[!htb]
			\centering
			\includegraphics[width=0.95\linewidth]{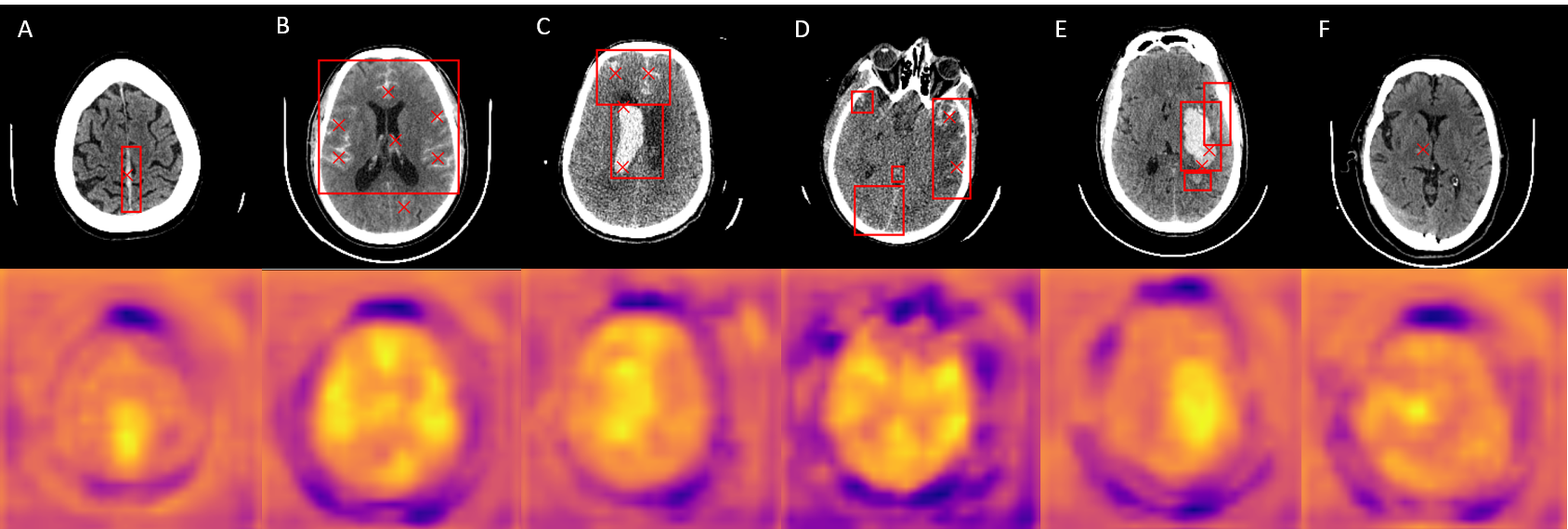}
			\caption{Top: Examples of ICH localization results (red $\times$-markers) together with annotated BB (red rectangle). Bottom: Generated attention maps; yellow color denotes to higher pixel values in contrast to blue. A -- Correct localization of single subdural hemorrhage. B -- Precise localization of individual subarachnoid bleedings with one FP finding. C -- Multiple intraventricular and subarachnoid hemorrhages localization resulting from multiple peaks in attention map. D -- Bleedings on the right are localized correctly; other bleedings are displayed in the attention map, however their maximal values are insufficient to be localized. E -- Multiple bleedings act like one massive high intensity area with two local maxima in the attention map, hence only ICH in the middle is localized. F -- False localization of unannotated high intensity region.}
		\label{fig:results}
\end{figure*}

\section{Results and discussion}

The proposed detection method was tested on the remaining 3/5 of data from the CQ500 dataset (i.e., 254 scans, more than 55,000 slices) using slice-level BB annotations. DC together with sensitivity (Se) and positive predictive value (PPV) were chosen to evaluate the algorithm. To compute the metrics, the total number of true positive (TP), false positive (FP), and false negative (FN) results were calculated. TP is defined as an intersection of the localized position and BB (e.g., Fig. \ref{fig:results} C shows 2 TPs); FP is the detection outside a BB (e.g., the isolated lower finding in Fig. \ref{fig:results} B); FNs are undetected BBs (e.g., Fig. \ref{fig:results} D shows 3 FNs). The overall results for the test dataset are shown in Tab. \ref{tab:results}. The validation classification accuracy (on the RSNA dataset) of the CNN was 89.54 \% and 91.33 \% for the model with max-pooling and attention layer, respectively.

\begin{table}[!htb]
\centering
\caption{The results of ICH localization algorithm on the test dataset and comparison of both methods that were used to generate likelihood maps: max-pooling and attention layer. PPV (positive predictive value) indicates probability, that localized position truly is ICH, SE (sensitivity) reflects the percentage of correct localizations out of all ICHs.}
\label{tab:results}
\begin{tabular}{c||c|c|c}
  \bfseries Method   & \bfseries PPV	[\%]	  &  \bfseries SE [\%]  &  \bfseries Dice [\%]  \\ \hline	\hline
Pooling    &  61.88  &  54.72  &  58.08  \\ \hline
 Attention  &  38.19   &  62.13   &  47.30  \\ 
\end{tabular}
\end{table}

In this study, we proposed a weakly supervised deep-learning-based ICH localization algorithm. Having only slice-level annotations, the detector was trained to predict the precise position of ICHs in axial CT slices, which is the main advantage of the proposed method. Classification CNN was used to predict likelihood maps giving information about possible areas of bleeding. Maps highlight the regions of CNN attention to predict the classification result. Hence, likelihood maps may be considered as the interpretation of the final classification. 

Precise ICH positions were found by peak detection of the likelihood map. The detector was optimized and tested on a publicly available dataset (different to training data) to test the generalization ability of the proposed approach to new data and to make the results comparable to other authors. According to the evaluation results, the attention maps given by the max-pooling layer seem more appropriate for the localization. The detector can find any type of ICH despite its large size, shape, and location variability. Besides, localization of multiple bleedings is possible. Even difficult-to-recognise bleedings are detected -- e.g., subdural hemorrhage in the higher intensity region of the cerebral falx (Fig. \ref{fig:results} A) or small bleeding such as little intraparenchymal or subarachnoid hemorrhages (Fig. \ref{fig:results} B, C, D).

FN results might occur in some cases of small IPH surrounded by large edema. Both FN and FP results sometimes originate in the detected positional coordinates closing aboard the BB. Besides, unannotated high-intensity regions subjectively similar to an ICH cause FP detections (Fig. \ref{fig:results} F).

Despite the possibility of false results, the detection ability denotes the high potential of the algorithm to minimize the probability of missing an ICH by oversight in clinical practice. Incorporation of the method in a computer-aided diagnostic system might warn a radiologist by highlighting the possible locations of ICHs while examining axial slices of a CT scan. Besides, the algorithm might significantly decrease the examination time as the processing of a slice takes only a few seconds. Considering the aforementioned arguments, the clinical use of the algorithm might help to prevent permanent disability or even death.

\section{CONCLUSIONS}
This paper demonstrates a fully automated, weakly supervised method for the localization of ICHs in axial head CT slices. The proposed algorithm is based on local extrema detection in attention maps that are obtained by a deep learning model. The main advantage of the algorithm is the proposed MIL position-free learning method, which is used for attention map generation. Hence, our approach showed the ability to localize ICHs using only slice-level annotations. The Dice coefficient of 58.08 \% was achieved on data from the publicly available dataset.




\section*{ACKNOWLEDGMENT}

Computational resources were supplied by the project "e-Infrastruktura CZ" (e-INFRA LM2018140) provided within the program Projects of Large Research, Development and Innovation Infrastructures.


\bibliographystyle{IEEEtran}
\bibliography{mybibfile}

\end{document}